# Efficient Online Surface Correction for Real-time Large-Scale 3D Reconstruction

Robert Maier*
maierr@in.tum.de

Raphael Schaller*
schaller@in.tum.de

Daniel Cremers
cremers@in.tum.de

Computer Vision Group
Technical University of Munich
Munich, Germany

* equal contribution

## Abstract

State-of-the-art methods for large-scale 3D reconstruction from RGB-D sensors usually reduce drift in camera tracking by globally optimizing the estimated camera poses in real-time without simultaneously updating the reconstructed surface on pose changes. We propose an efficient on-the-fly surface correction method for globally consistent dense 3D reconstruction of large-scale scenes. Our approach uses a dense Visual RGB-D SLAM system that estimates the camera motion in real-time on a CPU and refines it in a global pose graph optimization. Consecutive RGB-D frames are locally fused into keyframes, which are incorporated into a sparse voxel hashed Signed Distance Field (SDF) on the GPU. On pose graph updates, the SDF volume is corrected on-the-fly using a novel keyframe re-integration strategy with reduced GPU-host streaming. We demonstrate in an extensive quantitative evaluation that our method is up to 93% more runtime efficient compared to the state-of-the-art and requires significantly less memory, with only negligible loss of surface quality. Overall, our system requires only a single GPU and allows for real-time surface correction of large environments.

## 1 Introduction

In recent years, there has been a boost of research in the field of dense 3D reconstruction due to the wide availability of low-cost depth sensors such as the Microsoft Kinect. Most of the approaches fuse depth maps obtained from such sensors in real-time into a volumetric surface representation [3] to compensate for sensor noise and perform frame-to-model camera tracking against the fused volume. While researchers have shown the suitability of these methods for accurate geometric reconstruction of objects or scenes of limited size [16], global drift in camera tracking is not compensated, limiting the reconstruction of large-scale environments [8, 17, 19].

However, only few methods tackle the problem of globally optimizing the camera poses in real-time and simultaneously correcting the reconstructed surface on-the-fly. BundleFusion by Dai et al.[4] represents the state-of-the-art and estimates highly accurate camera poses on a high-end GPU. They require a second graphics card for integrating input RGB-D frames into a sparse Signed Distance Field (SDF) volume, making the entire framework computationally demanding. On pose graph updates, BundleFusion corrects the reconstructed

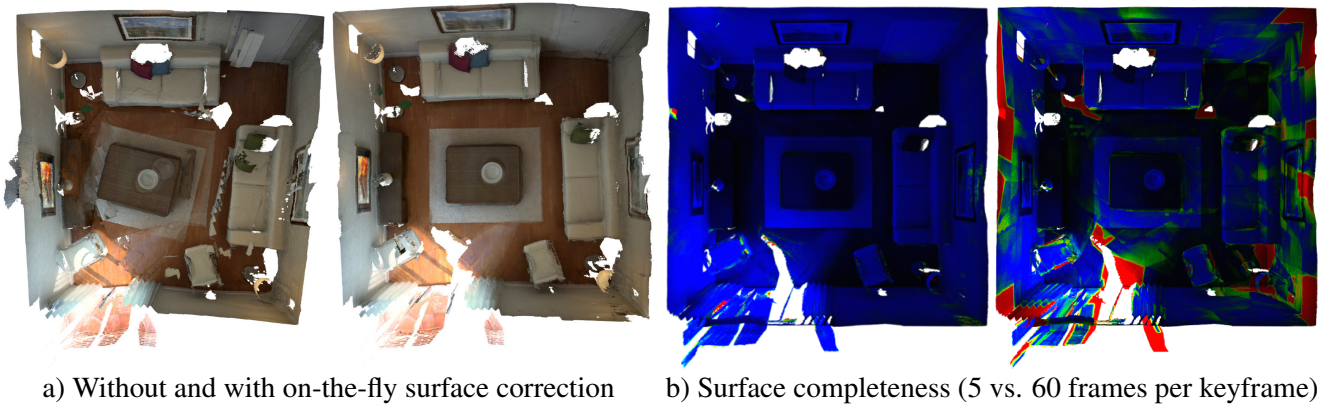

a) Without and with on-the-fly surface correction    b) Surface completeness (5 vs. 60 frames per keyframe)

Figure 1: Our method efficiently corrects the surface during the 3D scanning process on-the-fly (a) using an efficient keyframe re-integration strategy. Fusing fewer frames into each keyframe allows to maintain the completeness of the reconstructed 3D model (b).

surface on-the-fly by frame re-integration. However, all previous frames need to be held in memory to allow for a fast re-integration on pose updates; this limits its suitability for scanning large-scale environments with long sequences.

To enable state-of-the-art large-scale 3D reconstruction from RGB-D sensors, our SLAM framework is based on DVO-SLAM by Kerl et al. [10] for estimating a globally consistent camera motion. The system is computationally significantly less expensive than Bundle-Fusion and works in real-time on a single CPU, with only slightly less accurate estimated camera poses. To obtain globally consistent and up-to-date reconstructions of large environments, we couple it with our novel 3D surface correction method. Figure 1 shows the result of our online surface re-integration method at the end of a 3D scanning session and indicates the effect of keyframe fusion on the completeness of the reconstruction.

In summary, the main contributions of our work are:

- We integrate our 3D surface correction framework with a dense Visual SLAM system, such that our entire 3D reconstruction system runs in real-time with a single GPU only.
- We fuse consecutive RGB-D input frames in keyframes of high depth and color quality using different keyframe strategies.
- Our surface correction framework is highly efficient by only re-integrating fused keyframes into a sparse SDF volume on pose graph updates.
- Our strategy for selecting keyframes to be updated substantially reduces streaming between host and GPU.
- An extensive quantitative evaluation shows that our method is overall 93% more efficient compared to the state-of-the-art while maintaining surface quality.

## 2 Related Work

The field of dense 3D reconstruction from RGB-D data has been investigated extensively in recent years. KinectFusion by Newcombe et al. [16] enabled dense 3D reconstruction in real-time through extensive use of GPU programming. Like most of the following approaches, it stores the 3D model in an SDF volume [3], which regularizes the noisy depth maps from RGB-D sensors, and performs ICP-based camera tracking against the raycasted 3D model. Voxel Hashing [17] better exploits scarce GPU memory and allocates only occupied voxel blocks of the SDF. A hash map flexibly maps 3D voxel block coordinates onto memory locations. Kähler et al. [8] designed an optimized version of Voxel Hashing for mobile devices. However, the frame-to-model camera tracking of the frameworks above is only of limited use for reconstructing larger scenes. To reduce drift explicitly, recent approaches [1, 13, 18, 22] rely on loop closure detection in combination with global pose optimization.

In order to efficiently estimate camera poses in real-time, DVO-SLAM by Kerl et al. [10] minimizes a photometric and geometric error to accurately align RGB-D frames. For global consistency, it continuously performs a pose graph optimization to reduce global drift. While there is no dense volumetric model representation, they exploit keyframes to reduce the influence of noise. The system provides an excellent trade-off between runtime and accuracy, making it highly suitable for our 3D reconstruction framework. Utilizing keyframes as intermediate representation for reducing noise has also been exploited for improving camera tracking [15] and reconstruction appearance [14]. Following this idea, we also employ keyframes in our work as memory efficient intermediate 2.5D representations of 3D surfaces.

There are only few works on real-time large-scale RGB-D based 3D reconstruction that incorporate online surface correction. Fioraio et al. [5] reconstruct overlapping subvolumes, register their poses globally and update the subvolumes using volume blending. However, the absence of loop closure detection avoids to cope with larger drift. Kähler et al. [9] perform real-time tracking against multiple submaps independently and globally optimize the estimated trajectories. Submaps are fused on-the-fly during raycasting.

Whelan et al. use a deformation graph for online update of a surfel-based model [21] and of an SDF model [20] with an as-rigid-as-possible surface deformation. In ElasticFusion [21] input frames are fused into surfels and then discarded. However, wrong camera poses (e.g. due to drift) result in inconsistent surfel observations and hence increase their uncertainties; surfels with high uncertainty are ultimately filtered out. When a loop closure is detected, only the existing surface can be corrected along with the deformation graph, while surface information lost through inconsistent fusion cannot be recovered. In contrast, our method keeps all keyframes fused from input data and allows to re-integrate them at any pose graph update without a loss of surface information. Additionally, despite correcting the model online, the frame-to-model camera tracking may fail to compensate for drift due to delayed surface updates and undetected (or too late detected) loop closures.

BundleFusion et al. [4] represents the state-of-the-art both w.r.t. SLAM system accuracy as well as on-the-fly surface re-integration. The system first optimizes consecutive frame poses locally within chunks, which are then aligned globally in a hierarchical global optimization. New RGB-D input frames are matched brute-force against all previous chunk keyframes and subsequently aligned using a sparse-then-dense alignment. The global alignment regularly changes camera poses; to correct the reconstructed sparse SDF volume on-the-fly, the system first de-integrates frames with their former poses and then integrates them with their updated poses using a simple re-integration strategy. The 3D model is gradually adapted to the updated poses while still enabling real-time reconstruction. In contrast to BundleFusion, our method needs only a single GPU for surface modeling instead of two high-end graphics cards. Our online surface re-integration combines keyframe fusion with a more intelligent keyframe selection strategy, resulting in a significantly more efficient re-integration. Moreover, the use of keyframes requires substantially less memory and enables on-the-fly surface correction for large environments.

## 3  3D Reconstruction System

Our framework consists of a real-time RGB-D SLAM framework for globally consistent camera pose estimation and a sparse SDF volume for storing the reconstructed 3D model.

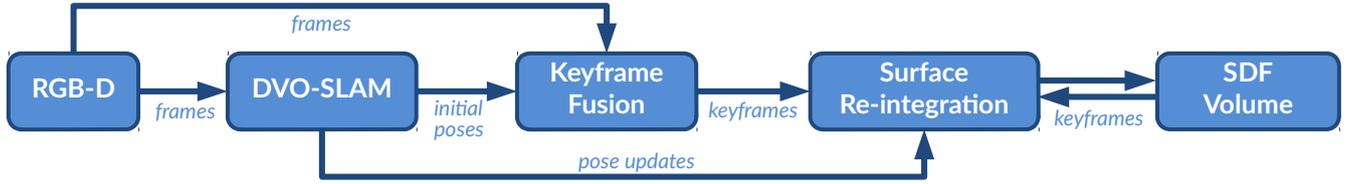

Figure 2: Overview of our online surface correction method. RGB-D frames are fused into keyframes, which are (re-)integrated into the SDF on-the-fly on DVO-SLAM pose updates.

While dense SDF-based 3D reconstruction methods usually integrate new RGB-D input frames directly into the volume, we first fuse them into keyframes as intermediate data representation. We integrate and re-integrate them online into the SDF on pose updates to efficiently correct the surface. This way, we can reduce the number of individual frames that we integrate into the SDF volume, which helps especially when we need to correct the 3D model due to a pose update. Figure 2 shows an overview of our approach.

**Preliminaries** We acquire RGB-D data from commodity depth sensors with 30 fps at a resolution of $640 \times 480$ pixels. The $N$ captured RGB-D frames consist of registered color images $\mathcal{C}_i$, depth maps $\mathcal{Z}_i$ and camera poses $\mathcal{T}_i = (R_i, t_i) \in \text{SE}(3)$ (with $R_i \in \text{SO}(3)$, $t_i \in \mathbb{R}^3$ and $i \in 1\ldots N$). A 3D point $p = (X,Y,Z)^\top$ is transformed using a pose $\mathcal{T}_i$ through $g(\mathcal{T}_i, p) = R_i p + t_i$. We use the pinhole camera model, which projects 3D points $p$ to 2D pixels $x = (u,v)^\top = \pi(p)$ using the projection $\pi : \mathbb{R}^3 \mapsto \mathbb{R}^2$. The inverse projection $\pi^{-1} : \mathbb{R}^2 \times \mathbb{R} \mapsto \mathbb{R}^3$ maps a 2D pixel location $x$ back to the respective 3D point $p = \pi^{-1}(x, \mathcal{Z}(x))$ using its depth.

**Dense Visual RGB-D SLAM** To estimate globally consistent camera poses $\mathcal{T}_i$, we utilize the DVO-SLAM system by Kerl et al. [10]. It runs in real-time on a CPU and employs a robust dense visual odometry approach that minimizes the photometric and geometric error of all pixels to estimate the rigid body motion between two RGB-D frames. To reduce drift in camera pose estimation, input frames are aligned against the preceding keyframe. Keyframes are selected using the differential entropy of the motion estimate and a pose distance threshold. In the following, we refer to the keyframes selected by DVO-SLAM as *DVO keyframes*. DVO-SLAM detects loop closures by aligning keyframes against candidates of previous keyframes within a sphere of predefined radius and validates them using their entropy ratio. Estimated frame-to-(key)frame camera motions and successful loop closures are integrated as constraints into a graph based map representation. This keyframe pose graph is steadily optimized in the background during runtime, yielding a globally consistent camera trajectory with continuously updated keyframe poses $\mathcal{T}_i$. Please note that our surface correction method works in principle with any SLAM system that incorporates loop closures.

**Keyframe Fusion** Our keyframe fusion builds up on [14] and consists of separate steps for depth and color fusion (cf. Figure 3). While new depth maps are immediately fused into the keyframe depth, color fusion relies on the more complete fused keyframe depth.

For depth fusion, we first compute for each pixel $x$ of $\mathcal{Z}_i$ its respective view- and distance-dependent weight $w_z(x) = \cos(\theta_i(x)) \cdot \mathcal{Z}_i(x)^{-2}$, where $\theta_i(x)$ is the angle between the depth normal at $x$ and the camera axis. Furthermore, we discard error-prone depth values close to depth discontinuities. We then warp each pixel with the frame pose $\mathcal{T}_i$ into the keyframe with pose $\mathcal{T}^*$ and obtain $p^* = (X^*, Y^*, Z^*)^\top = g(\mathcal{T}^{*-1}, g(\mathcal{T}_i, \pi^{-1}(x, \mathcal{Z}_i(x))))$. The keyframe depth $\mathcal{Z}^*$ and the depth fusion weights $\mathcal{W}^*$ at the projected 2D image point $x^* = \pi(p^*)$ are then updated as follows:

$$\mathcal{Z}^*(x^*) = \frac{\mathcal{W}^*(x^*)\mathcal{Z}^*(x^*) + w_z(x)Z^*}{\mathcal{W}^*(x^*) + w_z(x)}, \qquad \mathcal{W}^*(x^*) = \mathcal{W}^*(x^*) + w_z(x). \tag{1}$$

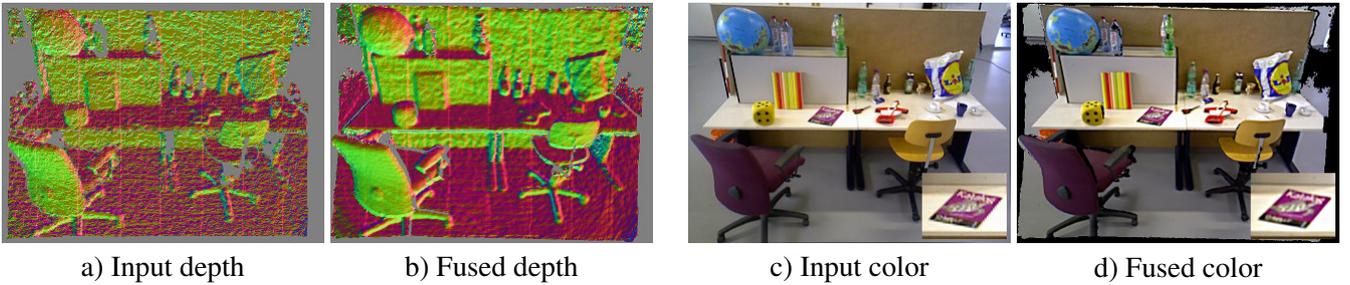

a) Input depth   b) Fused depth   c) Input color   d) Fused color

Figure 3: Keyframe fusion: several consecutive input depth maps (a) are fused into the keyframe depth (b). Our color fusion creates sharp color keyframes (d) from input color (c).

For color fusion, we first deblur the input color images using *Unsharp Masking* and compute a per-frame blurriness measure $w_b$ [2] from $\mathcal{C}_i$ to alleviate frames with strong motion blur. In contrast to depth fusion, the fused keyframes are warped back into each input frame $i$ and the observed color values $c_i(x^*) = \mathcal{C}_i(\pi(g(\mathcal{T}_i^{-1}, g(\mathcal{T}^*, \pi^{-1}(x^*, \mathcal{Z}^*(x^*))))))$ are sampled using bilinear interpolation. For a pixel $x^*$ in the keyframe, we collect all valid color observations in the input views and their color weights $w_{c_i}(x^*) = w_b \cdot w_z$. We compute the final keyframe color $\mathcal{C}^*(x^*)$ as the weighted median of the collected observations.

**SDF volume** At the core of our method, we store a memory efficient sparse SDF volume based on Voxel Hashing [17] as volumetric 3D model representation for large-scale 3D reconstructions. The implemented data structure is tailored to GPUs and only occupied space is allocated in voxel blocks, which are efficiently addressed using spatial hashing. For each voxel $v$, we store its signed distance $\mathbf{D}(v)$, its color $\mathbf{C}(v)$ and its integration weight $\mathbf{W}(v)$. We extract the iso-surface from the SDF using Marching Cubes [12]. To overcome the limitations of scarce GPU memory for large-scale environments, voxel blocks are streamed from GPU to host (and vice versa) before integration of a new frame. In particular, only voxel blocks within a sphere of constant radius around the current camera position are kept in GPU memory, while all other voxel blocks are streamed to the host. When an RGB-D frame is integrated into the SDF volume, voxel blocks are first allocated and the voxels are then updated using a running weighted average.

## 4 Efficient Online Surface Re-Integration

In the following, we introduce our online surface correction method that combines keyframe fusion with our sparse SDF volume implementation. Firstly, we incorporate on-the-fly keyframe re-integration into the 3D reconstruction pipeline; secondly, we show different strategies for starting new keyframes; thirdly, we propose an efficient surface correction procedure that is based on a re-integration strategy that reduces GPU-host transfer.

### 4.1 System Pipeline

While DVO-SLAM selects *DVO keyframes* for camera tracking based on an entropy criteria, we introduce *KF keyframes* (Keyframe Fusion keyframes) as intermediate representation for surface (re-)integration. When a new frame arrives, DVO-SLAM provides an initial pose estimate which is used to fuse the input frame into the current KF keyframe. Depending on the chosen keyframe selection strategy, a new keyframe will be started if some criteria are met and the previous KF keyframe is integrated into the SDF volume. The KF keyframe is also stored in memory for later re-integration on pose updates. Since DVO-SLAM issues only pose updates for DVO keyframes, we convert by expressing KF keyframe poses relative to DVO keyframe poses. The KF keyframes are then de-integrated from the SDF volume with their former camera poses and re-integrated on-the-fly with their updated poses.

| Frame | 1 | 2 | 3 | 4 | 5 | 6 | 7 | 8 | 9 | 10 | 11 | 12 | 13 | 14 | 15 |
|---|---|---|---|---|---|---|---|---|---|---|---|---|---|---|---|
| Distance | 1 | 3 | 4 | 3 | 5 | 4 | 1 | 7 | 2 | 1 | 1 | 8 | 6 | 2 | 0 |
| Distance (sum) | | | | 16 | 19 | 17 | **20** | 19 | 15 | 12 | 19 | 18 | 18 | 17 | |

Figure 4: Selection of frames for re-integration: BundleFusion [4] chooses the frames with highest distances between integrated pose and new pose. However, selecting frames 12, 8, 13, 5, 3 results here in disadvantageous shifts of the streaming sphere. Our method selects the group of most-moved $m$ consecutive frames, which results in frames 4 to 8 ($j^* = 4$).

## 4.2 Keyframe Strategies

In the following, we investigate keyframe selection strategies w.r.t. obtaining optimal surface quality. With only few input frames fused into KF keyframes, many KF keyframes need to be (re-)integrated into the SDF volume on pose updates. On the other hand, fusing many input frames into KF keyframes leads to a degradation in 3D reconstruction quality, since the 2.5D keyframes cannot fully represent the incorporated 3D information. We present keyframe strategies to find the optimal trade-off between re-integration performance and reconstruction quality.

The KF_CONST strategy is a simple but effective strategy and fuses a constant number $\kappa$ of frames into each KF keyframe. KF_DVO uses the frames selected as DVO keyframes also as KF keyframes. The distance based strategy KF_DIST issues a new KF keyframe whenever the rotational distance $\Delta_r$ or translational distance $\Delta_t$ of the relative pose $\mathcal{T}_{ij}$ between the current frame and the current KF keyframe exceeds a certain threshold, similar to [11]. The overlap strategy KF_OVRLP is derived from [7] and generates a new KF keyframe when the ratio of the pixels visible in both current frame and keyframe drops below a threshold.

## 4.3 On-the-fly Surface Correction

Our surface correction method follows the frame re-integration approach of [4]. However, we substantially improve it at critical points w.r.t. runtime efficiency by implementing a more intelligent strategy for selecting the KF keyframes to be re-integrated. Since we only need to correct KF keyframes instead of all frames, our surface correction is highly efficient w.r.t. runtime and memory consumption.

**Frame de-integration** For de-integrating an RGB-D frame $i$ from the SDF volume, we simply reverse the integration procedure. We therefore retrieve the KF keyframe from the memory and compute the projective distance $d_i$ (along the z axis) of $v$ in depth map $\mathcal{Z}_i$ (with sampling weight $w_i$) and its sampled color $c_i$ in the input color image $\mathcal{C}_i$. The de-integration steps for updating signed distance, color and weight of a voxel are denoted as follows:

$$\mathbf{D}'(v) = \frac{\mathbf{D}(v)\mathbf{W}(v) - d_i w_i}{\mathbf{W}(v) - w_i} \qquad \mathbf{C}'(v) = \frac{\mathbf{C}(v)\mathbf{W}(v) - c_i w_i}{\mathbf{W}(v) - w_i} \qquad \mathbf{W}'(v) = \mathbf{W}(v) - w_i \qquad (2)$$

**Re-integration strategy** While the poses of all keyframes are updated when DVO-SLAM issues a pose update, it is computationally too expensive to correct them all immediately. Instead, we re-integrate only $m$ changed frames whenever we receive a pose update. Poses that were not corrected on-the-fly are re-integrated in a final pass after the reconstruction. We denote the SDF integration pose of a frame by $\mathcal{T}_i$ and the updated pose by $\mathcal{T}_i'$.

To select the *m* frames for re-integration, BundleFusion orders all frames by descending distance between $\mathcal{T}_i$ and $\mathcal{T}_i'$ $\|st_i - st_i'\|$ and selects the *m* most-moved frames. The vectors $t_i$ and $t_i'$ contain the Euler rotation angles and the translation of the poses $\mathcal{T}_i$ and $\mathcal{T}_i'$, with a constant scale vector $s = (2,2,2,1,1,1)^\top$. However, since the corrected frames (and the respective SDF voxel blocks) may be spatially distant, suboptimal expensive GPU-host-streaming of voxel blocks may be required. To limit the streaming overhead, it is beneficial to correct close frames within the same re-integration procedure. We therefore keep the original temporal ordering of frames and select the group of most-moved *m consecutive* frames:

$$j^* = \underset{j \in [1, K-m+1]}{\arg\max} \sum_{i=j}^{j+m-1} \|st_i - st_i'\|, \quad (3)$$

where *K* is the total number of frames integrated so far. The resulting $j^*$ represents the first frame of our *m* consecutive frames thats need to be re-integrated. Figure 4 exemplifies the advantages of this procedure. Additionally, we adjust the streaming procedure of [17] to the re-integration process: We first stream in all voxel blocks inside the sphere around pose $\mathcal{T}_{j^*}$ to safely access them. Then, we successively de-integrate frames $[j^*, j^* + m - 1]$ with regular streaming. After de-integration, we stream in the sphere around the updated pose $\mathcal{T}_{j^*}'$ und successively re-integrate frames $[j^*, j^* + m - 1]$ using their updated poses with regular streaming. We finally stream the sphere back to the next integration pose.

## 5 Evaluation and Experimental Results

To demonstrate the effectiveness of our surface reconstruction algorithm, we provide a thorough quantitative evaluation w.r.t. runtime efficiency and surface accuracy. In particular, we analyze the effects of combining keyframe fusion with our surface re-integration method.

**Datasets** We use publicly available RGB-D datasets of large-scale scenes with loop closures that provide registered depth and color images as well as the respective camera poses. *AUG_ICL/Liv1* (noisy) [1] is a synthetic RGB-D sequence that is rendered from a modeled scene of a living room with realistic sensor noise. In addition to ground truth poses it also provides the ground truth 3D scene model that allows for a quantitative comparison of surface quality of reconstructed 3D models. *BundleFusion/apt0* [4] features a long camera trajectory of 8560 frames with poses estimated from BundleFusion.

**Surface evaluation methods and metrics** The evaluation procedure for comparing our reconstructed 3D models with synthetic ground truth is adapted from [6] and first extracts a 3D mesh $\mathcal{M}$ from the reconstructed SDF volume. We use CLOUDCOMPARE [1] to uniformly sample a reference point cloud $\mathcal{R}$ with 50 million points from the ground truth mesh of *AUG_ICL/Liv1*. We measure the distance of each vertex of $\mathcal{M}$ to its closest vertex in $\mathcal{R}$ with SURFREG [2] and compute the mean absolute deviation MAD. This technique assesses the *correctness* CORR of the model, i.e. the accuracy of the successfully reconstructed surfaces. However, we also want to measure the *completeness* COMPL of reconstructions to determine the information loss from keyframe fusion. For measuring COMPL, we inversely compare every vertex of $\mathcal{R}$ to the nearest neighbor in $\mathcal{M}$. For a fair comparison and to only compare surfaces visible in the synthetic frames, we re-generate the reference $\mathcal{R}$ by fusing all input frames into the SDF with ground truth poses. We rely on the poses from the datasets for assessing the surface quality to eliminate a substantial source of error. We used a workstation with Intel Core i7-3770 CPU, 32GB RAM and an NVIDIA GeForce GTX 1070 GPU.

---
[1] http://www.danielgm.net/cc/
[2] https://github.com/mp3guy/SurfReg

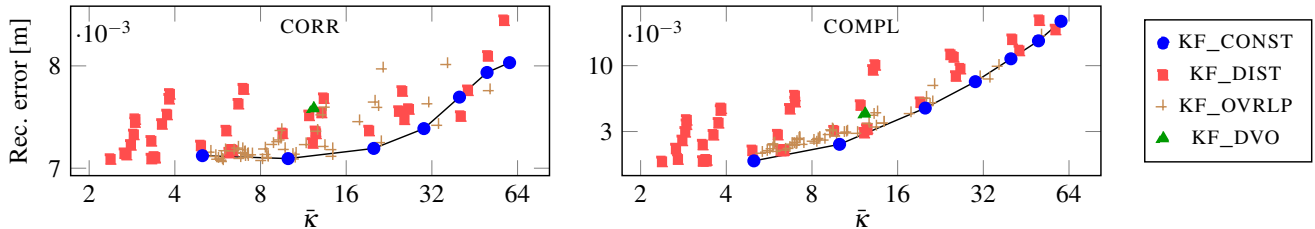

Figure 5: Quantitative evaluation of reconstruction correctness (left) and completeness (right) w.r.t. different keyframe strategies on *AUG_ICL/Liv1*. The *x*-axis shows the average keyframe size $\bar{\kappa}$ produced in each run, the *y*-axis shows the MAD error (axes are logarithmic). The KF_CONST strategy achieves the best reconstruction results for both CORR and COMPL.

## 5.1 Keyframe Fusion

We quantitatively investigate the effect of keyframe fusion on the reconstruction quality, i.e. surface completeness and correctness, of the noisy *AUG_ICL/Liv1* dataset.

**Keyframe strategies** Figure 5 shows the results of different keyframe selection strategies and their average keyframe sizes on the reconstructed surface quality. Each mark represents a separate evaluation run of a given strategy with a different set of specified parameters. For KF_CONST, we vary the number of consecutive frames $\kappa$ that are fused into each keyframe. In KF_DIST we adjusted the pose distance threshold $\Delta_t$ and $\Delta_r$, while we varied the overlap ratio parameters in KF_OVRLP. In KF_DVO we use the same keyframes as DVO-SLAM. In summary, more relaxed parameters result in a higher average number of fused frames per keyframe $\bar{\kappa}$ for all strategies; different parameter combinations for the same strategy may result in a similar $\bar{\kappa}$. With an increasing $\bar{\kappa}$, the completeness of reconstructions decreases rapidly, since 3D surface information gets lost in 2.5D keyframe fusion. The effect on surface correctness is less significant, since the deviation for the remaining surfaces is still reasonably close to the ground truth 3D model. Compared to KF_CONST, the strategies KF_DIST, KF_OVRLP and KF_DVO result mostly in worse quantitative results, a hardly predictable number of keyframes and barely tunable interdependent parameters. We found KF_CONST to give good quantitative results, while it is also highly predictable w.r.t. fusion and re-integration runtime as well as memory consumption ($\sim 1/\bar{\kappa}$) due to its priorly known number of frames per keyframe. We refer the reader to the supplementary material for more details.

**Completeness** Figure 6 shows color coded distance renderings for KF_CONST keyframe fusion with $\kappa = 5$ and $\kappa = 60$ on *AUG_ICL/Liv1*. The colors represent errors from 0mm (blue) to 50mm (red). Again, the completeness COMPL of reconstructions decreases with more fused frames per keyframe because of the loss of surface information with 2.5D keyframes. The reconstructed surfaces are still accurate w.r.t. ground truth (CORR).

## 5.2 Surface Re-integration

We finally assess our surface correction w.r.t. real-time performance and show results of on-the-fly surface re-integration on real-world data.

**Runtime** While BundleFusion requires two high-end GPUs to operate in real-time, our system requires only a single GPU. Figure 7 gives the average amortized runtimes of our system, specifically for (re-)integration of frames w.r.t. $\kappa$ (red), for DVO-SLAM (green) and

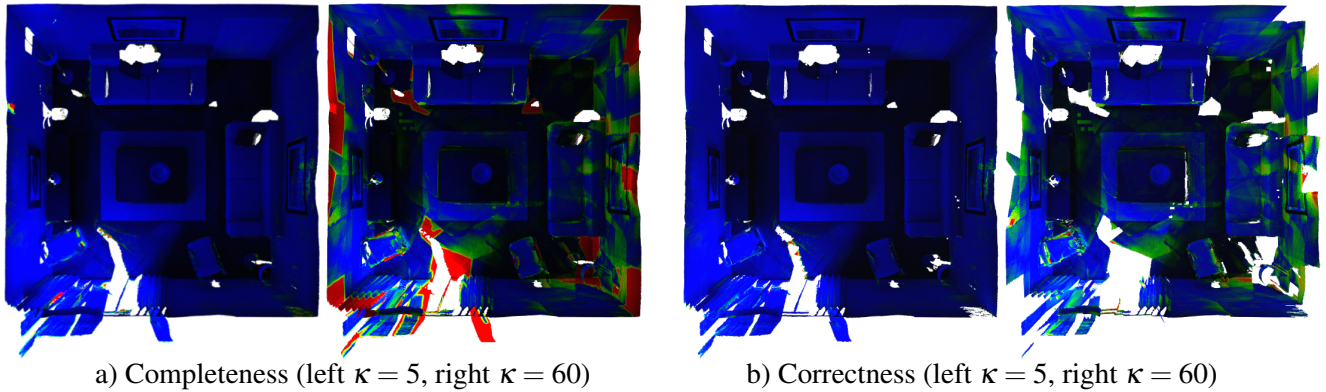

a) Completeness (left $\kappa = 5$, right $\kappa = 60$)   b) Correctness (left $\kappa = 5$, right $\kappa = 60$)

Figure 6: Completeness and correctness after integration of *AUG_ICL/Liv1* with KF_CONST keyframe strategy (with keyframe sizes 5 and 60).

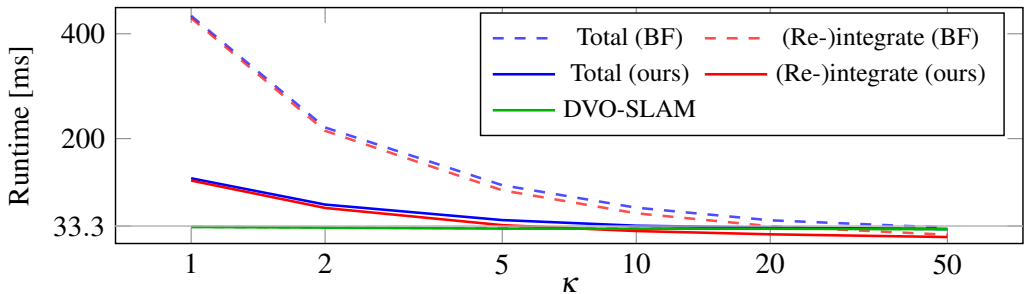

Figure 7: Average runtime per frame for reconstruction of *AUG_ICL/Liv1* with KF_CONST w.r.t. $\kappa$. Our re-integration strategy (solid) is substantially faster than BundleFusion's (dashed). $m$ was set to $100/\kappa$, yielding a constant effective re-integration rate.

the total runtime of both combined (blue). Generally, the higher $\kappa$ is, the fewer keyframes are generated and thus need to be updated. We accomplish real-time performance with $\kappa = 20, m = 5$. Here, the re-integration strategy of updating the $m$ most-moved *consecutive* keyframes (solid lines) saves already 47% of (re-)integration runtime compared to BundleFusion's simple strategy (dashed). This is further accelerated through the use of keyframes only: Overall, the reconstruction with $\kappa = 20, m = 5$ and our re-integration strategy takes 93% less time than BundleFusion's re-integration strategy without keyframe fusion ($\kappa = 1, m = 100$). We found $m \in [10, 20]$ to be a good trade-off between reconstruction quality and model correction speed for most data sets.

**On-the-fly surface re-integration** As DVO-SLAM steadily optimizes a pose graph and issues pose updates, our surface correction method gradually improves the reconstructed 3D model on-the-fly by re-integrating the most-moved consecutive $m$ keyframes into the SDF. While updating all changed keyframes at once is too expensive, we can control the speed of incorporating pose updates into the 3D model by adjusting $m$. Also, with decreasing $\bar{\kappa}$ more keyframes are generated and need to be updated. Figure 8 shows an example of how a 3D reconstruction is corrected on-the-fly during the reconstruction to be as globally consistent as possible (with $m = 5$, $\kappa = 20$ and KF_CONST strategy).

An isolated comparison of the surface correction of ElasticFusion [21] with our method is not applicable since the respective SLAM systems may result in different camera trajectories. Nevertheless, Figure 9 shows a qualitative comparison of the generated models, which is in accordance with the findings in [4]. While ElasticFusion might benefit from camera tracking against the corrected model, the point cloud reconstructed with ElasticFusion with default parameters exhibits double walls and artifacts due to potentially undetected loop closures and surface warping artifacts. These effects are mitigated in the continuous surface mesh reconstructed from our method, which successfully corrects the model on-the-fly.

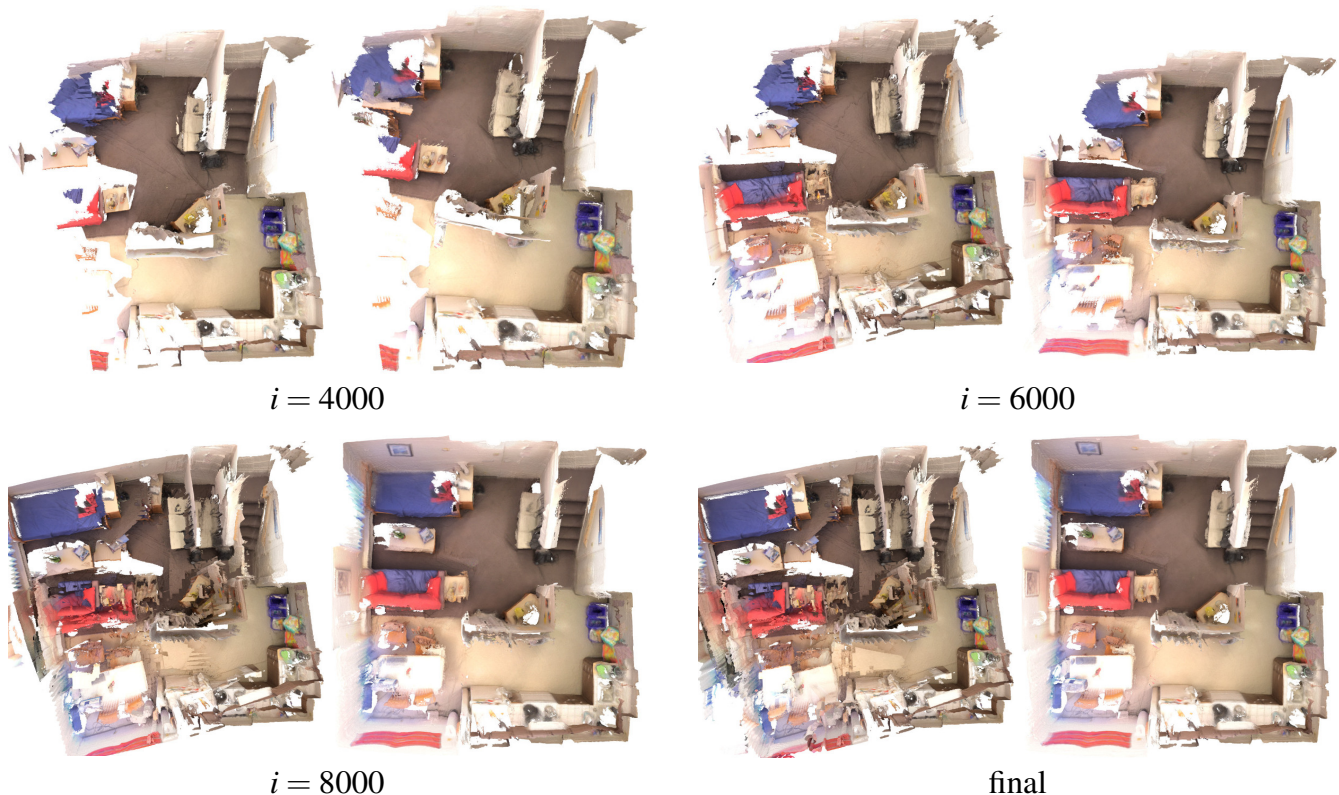

Figure 8: Reconstruction of *BundleFusion/apt0*. Every 2000 frames, a model was generated without (left) and with (right) on-the-fly surface correction (KF_CONST keyframe strategy with $m = 5$, $\kappa = 20$).

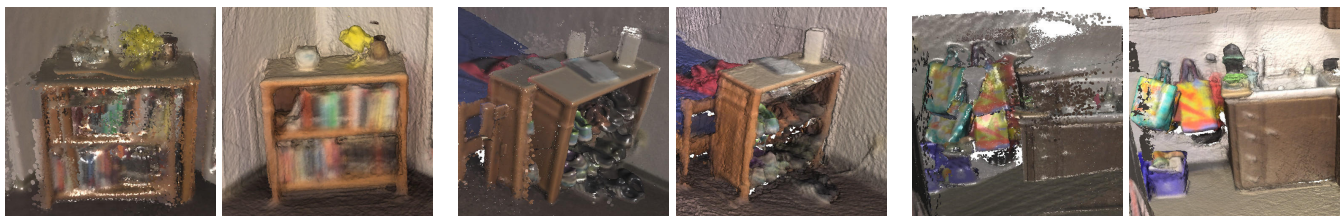

Figure 9: Qualitative comparison of ElasticFusion [21] (left) with our method (right) on *BundleFusion/apt0*. The point cloud reconstructed with ElasticFusion exhibits artifacts due to potentially undetected loop closures and surface warping artifacts, whereas our method successfully corrects the model.

## 6　Conclusion

We presented an efficient online surface re-integration method for globally consistent 3D reconstruction of large-scale scenes from RGB-D sensors in real-time on a single GPU only. Our SLAM system based on DVO-SLAM estimates the camera motion in real-time on a CPU and employs pose graph optimization for obtaining globally optimized camera poses. Multiple RGB-D frames are first fused into keyframes, which are then integrated into a sparse voxel hashed SDF model representation. Continuous keyframe pose updates are gradually incorporated into the SDF volume by on-the-fly re-integration of changed keyframes. Our improved re-integration strategy with correction of keyframes and significantly reduced host-GPU-streaming saves about 93% of runtime compared to the state-of-the-art. By re-integrating keyframes (instead of all frames), we substantially reduce the number of frames to be re-integrated with only a slight degradatation of reconstruction quality.

**Acknowledgement**　　This work was funded by the ERC Consolidator grant *3D Reloaded*.